\documentclass[journal,twoside,web]{ieeecolor}
\usepackage{tmi}
\usepackage{cite}
\usepackage{amsmath,amssymb,amsfonts}
\usepackage{algorithmic}
\usepackage{graphicx}
\usepackage{textcomp}
\usepackage{etoolbox}
\makeatletter
\@ifundefined{color@begingroup}%
  {\let\color@begingroup\relax
   \let\color@endgroup\relax}{}%
\def\fix@ieeecolor@hbox#1{%
  \hbox{\color@begingroup#1\color@endgroup}}
\patchcmd\@makecaption{\hbox}{\fix@ieeecolor@hbox}{}{\FAILED}
\patchcmd\@makecaption{\hbox}{\fix@ieeecolor@hbox}{}{\FAILED}
\usepackage{booktabs}
\usepackage{bbding}
\usepackage{multirow}
\usepackage{pifont}
\newcommand{\xmark}{\ding{55}}%
\input alphabet
\definecolor{mypurple}{rgb}{0.8,0.5,0.8}
\definecolor{myviolet}{rgb}{0.6,0.3,0.4}
\definecolor{cadmiumgreen}{rgb}{0.0, 0.42, 0.24}
\newcommand{\YZ}[1]{\textcolor{black}{#1}} 
\newcommand{\CH}[1]{\textcolor{black}{#1}}
\newcommand{\DM}[1]{\textcolor{black}{#1}} 
\newcommand{\DML}[1]{\textcolor{black}{#1}} 

\usepackage{soul}

\def\Diag{\mathrm{Diag}}

\def\R{\mathbb{R}}
\def\T{^\mathsf {T}}

\def\BibTeX{{\rm B\kern-.05em{\sc i\kern-.025em b}\kern-.08em
    T\kern-.1667em\lower.7ex\hbox{E}\kern-.125emX}}
\begin{document}
\bstctlcite{IEEEexample:BSTcontrol}  
\title{Diffusion Reconstruction of Ultrasound Images with Informative Uncertainty}

\author{Yuxin Zhang, Clément Huneau, Jérôme Idier, and Diana Mateus
\thanks{Manuscript received October 1, 2023.
This work has been supported by the European Regional Development
Fund (FEDER), the Pays de la Loire region on the Connect Talent scheme (MILCOM Project) and Nantes Métropole (Convention 2017-10470).
}
\thanks{Yuxin Zhang, Clément Huneau, Jérôme Idier, and Diana Mateus
are with Nantes Université, École Centrale Nantes, Laboratoire des Sciences du Numérique de Nantes LS2N, 
CNRS, UMR 6004, F-44000 Nantes, France.
(Corresponding author: yuxin.zhang@ls2n.fr).}
\thanks{This work has been submitted to the IEEE for possible publication. Copyright may be transferred without notice, after which this version may no longer be accessible.}
}

\maketitle

\begin{abstract}
Despite its wide use in medicine, ultrasound imaging faces several challenges related to its poor signal-to-noise ratio and several sources of noise and artefacts. Enhancing ultrasound image quality involves balancing concurrent factors like contrast, resolution, and speckle preservation. In recent years, there has been progress both in model-based and learning-based approaches to improve ultrasound image reconstruction. Bringing the best from both worlds, we propose a hybrid approach leveraging advances in diffusion models. To this end, we adapt Denoising Diffusion Restoration Models (DDRM) to incorporate ultrasound physics through a linear direct model and an unsupervised fine-tuning of the prior diffusion model. We conduct comprehensive experiments on simulated, in-vitro, and in-vivo data, demonstrating the efficacy of our approach in achieving high-quality image reconstructions from a single plane wave input and in comparison to state-of-the-art methods. Finally, given the stochastic nature of the method, we analyse in depth the statistical properties of single and multiple-sample reconstructions, experimentally show the informativeness of their variance, and provide an empirical model relating this behaviour to speckle noise. \textcolor{black}{The code and data are available at: (upon acceptance).}
\end{abstract}


\section{Introduction}
\label{sec:introduction}






\IEEEPARstart{U}ltrasound (US) \DML{imaging finds extensive use in musculoskeletal, cardiac, obstetrical, and other medical diagnostic applications. In contrast to MR or CT imaging, which are expensive or ionizing,  ultrasound is real-time, affordable, portable, and minimally invasive. However, US image quality is affected by acoustic artefacts (shadowing, reverberation, clutter), electronic and speckle noise and attenuation. Resolution and contrast are particularly impacted with unfocused transmissions, such as Plane Wave imaging, calling for dedicated digital reconstruction algorithms.}
%
\DML{While standard beamforming algorithms rely on the Delay-and-Sum (DAS)~\cite{DAS}  to transform raw signals into B-mode images, such low-complexity models lead to suboptimal image quality in terms of  signal-to-noise ratio (SNR), contrast, and spatial resolution, especially with unfocused US.}


Recent techniques to improve US imaging include \YZ{time-domain} adaptive beamforming techniques, e.g. \YZ{ Eigenspace-based Minimum Variance (EMV)~\cite{asl_eigenspace-based_2010} and Phase Coherence Imaging (PCF)~\cite{PCF}}, or Fourier-based reconstructions~\cite{Chernyakova-Eldar_2018}. Other methods focus on optimizing either \YZ{pre-beamforming raw signals~\cite{REFOCUS,khan_real-time_2021} or post-beamforming images~\cite{laroche_fast_2021}.}
Today, there is an increasing interest in model-based approaches that better formalize the \YZ{image reconstruction} problem within an optimization framework. For instance, 
\DML{
Ozkan \textit{et al.}~\cite{IPB_Ozkan} solve the inverse ultrasound reconstruction problem imposing $l_1$, $l_2$, or wavelet-based regularization terms  while Goudarzi \textit{et al.}~\cite{RED_USIPB} propose a
regularization by denoising (RED) approach.}
 
 A second branch of methods for improving US image quality leverages the power of Deep Neural Networks (DNNs). Certain techniques, like MobileNetV2 (MNV2)~\cite{MNV2} and Adaptive Ultrasound Beamforming using Deep Learning (ABLE)~\cite{luijten_adaptive_2020}, predict beamforming \DML{coefficients to weight} 
 the channel data, whereas 
 \cite{Hyun19,perdios_cnn-based_2022} 
 \DML{directly learn to predict high-quality B-mode images from channel or DAS beamformed data, respectively}. Despite their effectiveness, the above methods 
 \DML{have limited}
 interpretability and necessitate large amounts of low-high quality paired data for training. Such datasets are frequently scarce, especially within the medical domain.


\DML{Combining the best of the model-based and learning worlds, hybrid approaches focus on 
improving interpretability or 
removing the need of low-high quality training datasets}
For instance, Chennakeshava \textit{et al.}\ \cite{Chennakeshava:ius2020} solve a model-based plane-wave compounding problem by unfolding a classical optimization algorithm. 
Zhang \textit{et al.}~\cite{ZHANG2021} propose a self-supervised beamforming approach 
\DML{enforcing explicit prior assumptions on the reconstruction through the loss function.}
Our work falls within this hybrid model-based deep learning family of approaches \cite{van_sloun_deep_2020}, \YZ{harnessing the benefits of interpretability and a reduced reliance on extensive training datasets.}


In practice, we leverage the recent success of Denoising Diffusion Probabilistic Models (DDPMs) \cite{ho_denoising_2020,nichol_improved_2021,dhariwal_diffusion_2021},  which are the state-of-the-art in image synthesis. 
More specifically, we build on the Denoising Diffusion Restoration Models (DDRMs) framework proposed by Kawar et al.~\cite{kawar_denoising_2022}, which adapts DDPMs to various image restoration tasks modeled as linear inverse problems. The main advantage of DDRMs is exploiting the direct problem modeling to bypass the need to retrain DDPMs when addressing new tasks. 
\DML{We adapt DDRM to incorporathe the physics and constraints of ultrasound imaging through an approximate direct model, and with a self-supervised fine-tuning of the unconditional diffusion model.}

While the combination of model-based and diffusion models has been explored in the context of CT/MRI imaging \cite{song_solving_2022,chung2022scoreMRI}, 
\DML{these type of approach is new to ultrasound imaging.}
\DM{Two contemporary approaches have proposed alternative means to exploit diffusion models towards improving ultrasound imaging. Asgariandehkordi \textit{et al.} \cite{DenoDDPM} focus on denoising already beamformed images with a fine-tuned DDPM. In contrast, we define the direct and inverse problems linking raw data to US images and follow DDRM to address a reconstruction instead of a denoising problem. Our experimental validation shows our formulation leads to improved results over pure denoising. Like us, Stevens et al.\cite{dehaze} define the problem on the channel data but focus on the specific issue of dehazing for cardiovascular ultrasound images. To this end, \cite{dehaze} defines a joint posterior sampling framework combining two diffusion models independently representing clean data and the haze.
Instead we target the  general reconstruction problem and rely on a single diffusion model.
}

\YZ{The uncertainty of generative models has raised concerns regarding the reliability of the generated results \cite{horwitz2022conffusion}, particularly within the realm of medical imaging \cite{chung2022scoreMRI,dehaze}. Therefore, we conduct an analysis of the statistical properties of 
multiple samples and observed that their variance 
results in a noteworthy enhancement of the Signal-To-Noise Ratio (SNR) and contrast, while maintaining the resolution.} 

The contributions of our work are:

\DM{i)} An adaptation of DDRMs from restoration tasks in the context of natural images (e.g.\ denoising, inpainting, superresolution), to the reconstruction of B-mode US images from raw radio-frequency (RF) channel data. Our experiments focus on reconstructing an US image from a single plane wave. However, the proposed approach can be applied to different acquisition types, e.g. sequential imaging, synthetic aperture, and plane-wave, as long as the acquisition can be approximately modeled as a linear inverse problem, i.e.\ with a model matrix depending only on the geometry and pulse-echo response (point spread function). 

%
ii) An extensive quantitative evaluation on the PICMUS \YZ{dataset and a self-acquired \textit{in vivo} dataset} in the context of single plane wave imaging, which shows performance improvements when comparing against traditional DAS, state-of-the-art beamforming approaches \cite{asl_eigenspace-based_2010, PCF, RED_USIPB, MNV2, luijten_adaptive_2020, ZHANG2021}, and versus the recent proposition of using diffusion models as denoisers~\cite{DenoDDPM}.

%
iii) \YZ{Revealing, for the first time, that computing the variance of multiple samples of DRUS can achieve despeckling and result in an image with higher SNR and contrast.}

In addition to a full description of the direct and inverse problem, contributions ii) and iii) are extensions of our preliminary work \cite{zhang2023}.

\section{METHODS}
\DM{The main objective of this paper is to reconstruct an image that faithfully represents the reflectivity of the observed organs or objects from raw data acquired with an ultrasound probe, with a focus on improving single plane-wave imaging. Our approach is rooted in a standard direct linear model relating the image to the measurements (Section \ref{sec:forwardModel}). In contrast to conventional beamforming techniques or iterative algorithms addressing the inverse problem (Section \ref{sec:DAS}), we leverage a diffusion model to guide the image reconstruction. Specifically, we rely on a recent approach coupling inverse problems with diffusion models (Section \ref{sec:DDRM}). We further adapt this method to our computational constraints (Section \ref{sec:DRUS}) and provide an experimental analysis of its statistical properties.}

\subsection{Forward Ultrasound Imaging Model}
\label{sec:forwardModel}

For a given object to image, we denote $\ov$ its reflectivity map, which is the local variation of acoustic impedance. Imaging is performed using a transducer array of $L$ elements that emit a pulse $h_e$ and receive an echo signal $y$. In order to model the ultrasonic transmission-reception process as a linear \CH{time-}invariant system, we consider: the first-order Born approximation, a small variation of acoustic impedance and no absorption within the object.

The US wave emitted by the $i^{th}$ element passes through the \CH{object domain} $\Omega$ and is received by the $j^{th}$ element. The signal received at time $t$ can be expressed considering the propagation time $\tau_{i,j}$ for any position $\rv\in\Omega$ as

\begin{equation}
    y_{i, j}(t) = \int_{\rv \in \Omega}h(t-\tau_{i,j}(\rv)) o(\rv) \mathrm{d} \rv + n_{j}(t),
\label{Equ: model_continuous}
\end{equation}
where $n_{j}$ represents the noise for the $j^{th}$ receiving element, $h$ is \DM{a kernel resulting from} the convolution of the emitted excitation pulse $h_e$ and the transducer impulse response $h_t$.

In plane-wave ultrasound imaging, all elements in the transducer array emit the same pulse with a linear delay law, producing a plane wave that deviates from the normal of the transducer array with an angle $\alpha$. For a normal propagation direction ($\alpha=0$), the discretized model with $N$ observation position\DM{s} and $K$ time samples for all $L$ receiving elements can then be written as 
\begin{equation}
    \yv=\Hv\ov + \nv,
    \label{Equ: model_discret}
\end{equation}
where $\yv=[\yv_1\T,...,\yv_L\T]\T\in \R^{KL\times 1}$, $\ov\in \R^{N\times 1}$, $\nv=(\nv_1\T,...,\nv_L\T)\T\in \R^{KL\times 1}$, and $\Hv\in \R^{KL\times N}$. 
Matrix $\Hv$ is
\begin{equation}
    \Hv =
    \begin{bmatrix}
        \hv_{1,1} & \cdots & \hv_{j,N} \\
        \vdots & \vdots & \vdots \\
        \hv_{L,1} & \cdots & \hv_{L,N}  \\
    \end{bmatrix}_{KL \times N}
    \label{Equ: matrixH}
\end{equation}
with $\hv_{j,n}=\left [h(t_k-\tau_j(\rv_n))\right ]_{k\in[1:K]}\T$ a vector containing the kernel delayed for the $n$-th position to $j$-th element propagation and sampled at time $t_k$. Notably, $\Hv$ is a sparse matrix since $\hv_{j,n}$ is non-zero only when $t_k$ is close to $\tau_{j}(\rv_n)$.


Due to the linearity and invariance hypotheses, the inaccuracy of $h$ and the discretization, the additive noise $\nv$ does not only include white Gaussian electronic noise but also the model error. However, for simplicity, we still assume $\nv$ as white Gaussian with standard deviation $\gamma$, which is reasonable for the plane wave transmission~\cite{iMAP}.

\subsection{From Classical Beamforming to Inverse Problems}
\label{sec:DAS}
A common method to form an image $\hat\ov$ from received data $\yv$ is the DAS algorithm~\cite{DAS}. For each pixel position $\rv_n$, DAS sums the received data at the corresponding time-of-flight:
\begin{equation}
    \hat o_\mathrm{DAS}(\rv_n) = \sum_{j=1}^L y_j(\tau_j(\rv_n)),
\end{equation}
where the discretization of channel data implies the interpolation of one or more time samples around the exact time of flight. Further works~\cite{quaegebeur_correlation-based_2012} generalize DAS to a matched filtering,  calculated by:
\begin{equation}
    \hat \ov_\mathrm{DAS}=\Hv\T \yv.
\end{equation}
The practical implementation of DAS also considers an apodization to mitigate the limited directivity of elements of the transducer array~\cite{DAS}. Apodization factors $a_{j,n}$ between each element and pixel depend 
\DM{on the definition of a window's shape and aperture}
(the so-called \textit{f-number}). Apodization produces a weighted matched filter matrix
\begin{equation}
    \Bv =
    \begin{bmatrix}
        a_{1,1}\hv_{1,1} & \cdots & a_{1,N}\hv_{j,N} \\
        \vdots & \vdots & \vdots \\
        a_{L,1}\hv_{L,1} & \cdots & a_{L,N}\hv_{L,N}  \\
    \end{bmatrix}\T_{N\times KL}
    \label{Equ: matrixB}
\end{equation}
Finally, DAS beamforming can be expressed as:
%
\begin{equation}
    \hat \ov_\mathrm{DAS}=\Bv \yv.
\end{equation}
DAS-like beamforming methods 
are widely used because of their simple implementation and low calculation cost. 


\subsection{Diffusion Restoration Models}
\label{sec:DDRM}

A DDPM is a parameterized Markov chain trained to generate synthetic images from noise relying on variational inference~\cite{ho_denoising_2020,nichol_improved_2021,dhariwal_diffusion_2021}. 
The Markov chain consists of two processes: a forward fixed diffusion process and a backward learned generation process. Intuitively, the forward process gradually adds Gaussian noise with variance $\sigma_t^2$ ($t = 1,\ldots, T$) to the clean signal $\xv_0$ until it becomes random noise, while in the backward generation process, 
 the random noise $\xv_T$ undergoes a gradual denoising 
 until a clean $\xv_0$ is generated.
%

An interesting question in model-based deep learning is how to use prior knowledge learned by generative models to solve inverse problems. 
Denoising Diffusion Restoration Models (DDRM)~\cite{kawar_denoising_2022} were recently introduced for solving linear inverse problems, taking advantage of a pre-trained DDPM model as the learned prior. Similar to a DDPM, a DDRM is also a Markov Chain but conditioned on measurements $\yv_d$ through a linear observation model $\Hv_d$  
\footnote{We use subscript $d$ to refer to the original equations of the DDRM model.}. The linear model serves as a link between an unconditioned image generator and any restoration task. In this way, DDRM makes it possible to exploit pre-trained DDPM models whose weights are assumed to generalize over tasks. In this sense, DDRM is fundamentally different from previous task-specific learning paradigms requiring training with paired datasets. Relying on this principle, the original DDRM paper was shown to work on several natural image restoration tasks such as denoising, inpainting, and colorization.

Different from DDPMs, the Markov chain in DDRM is defined in the spectral space of the degradation operator $\Hv_d$. To this end, DDRM leverages the Singular Value Decomposition (SVD): 
$\Hv_d=\Uv_d \Sb_d \Vv_d\T$ with $\Sb_d=\Diag\left(s_1,\ldots,s_N\right)$,
which allows decoupling the dependencies between the measurements. 
The original observation model 
$
 \yv_d= \Hv_d \xv_d + \nv_d= \Uv_d \Sb_d \Vv\T _d\xv_d + \nv_d,
$
can thus be cast as a denoising problem that can be addressed on the transformed measurements:
\begin{equation*}
\overline{\yv}_d= \overline{\xv}_d + \overline{\nv}_d
\end{equation*}
with $\overline{\yv}_d= \Sb_d^\dag\Uv_d\T\yv_d$, $ \overline{\xv}_d=\Vv_d\T\xv_d$, and $\overline{\nv}_d=\Sb_d^\dag\Uv_d\T \nv_d$, where $\Sb_d^\dag$ is the generalized inverse of $\Sb_d$. The additive noise $\nv_d$ being assumed \textit{i.i.d.} Gaussian: $\nv_d\sim \mathcal{N}\left(0, \sigma_d^2\Iv_N\right)$, with a known variance $\sigma_d^2$ and $\Iv_N$ the $N\times N$ identity matrix, we then have
$\overline{\nv}_d$ with standard deviation $\sigma_d\Sb_d^\dag$.

Each denoising step from $\overline{\xv}_t$ to $\overline{\xv}_{t-1}$ ($t=T,...,1$) is a linear combination of $\overline{\xv}_t$, the transformed measurements $\overline{\yv}_d$, the transformed prediction of 
$\xv_0$ at the current step $\overline{\xv}_{\theta,t}$, and random noise. To determine their coefficients which are denoted as $A$, $B$, $C$, and $D$ respectively, the condition on the noise, $(A\sigma_t)^2 + (B\sigma_d/s_i)^2 + D^2 = {\sigma_{t-1}}^2$, and on the signal, $A+B+C = 1$, are leveraged, and the two degrees of freedom are taken care of by two hyperparameters.

In this way, the iterative restoration is achieved by the iterative denoising, and the final restored image is $\xv_0 = \Vv_d\overline{\xv}_{0}$. We denote the number of iterations as \textit{it}.

\subsection{Diffusion Reconstruction of Ultrasound Images}
\label{sec:DRUS}

Given the linear model~\eqref{Equ: model_discret}, we can now rely on DDRM to iteratively guide the reconstruction of an US image from the measurements. However, \CH{DDRM relies on the SVD of $\Hv$ to go from a generic inverse problem to a denoising/inpainting problem. An SVD requires storing huge orthogonal matrices that cannot be implemented as operators since $KL$ is always larger than $N$ in an ultrasound problem. In order to reduce the problem dimension we project the data}
using the beamforming matrix \DM{$\Bv$ (Section~\ref{sec:DAS})}:
\begin{equation}
    \Bv\yv=\Bv\Hv\ov + \Bv\nv,
    \label{Equ: model_HtH}
\end{equation}
In this way, the size of the SVD $\Bv\Hv=\Uv\Sb\Vv\T$ becomes more tractable. We then feed to DDRM the new model:
\begin{equation}
    \overline{\yv}= \overline{\ov} + \overline{\nv}
\end{equation}
to iteratively reconstruct $\overline{\ov}=\Vv\T\ov$ from $\overline{\yv}=\Sb^\dag\Uv\T\Bv\yv$ observations. \DM{In practice, a new image is obtained by DDRM sampling from a diffusion model (pretrained on natural images and fine-tuned on ultrasound) and relating the samples to the measurements \YZ{in \eqref{Equ: model_HtH}.} 
We name the resultant approach \textbf{\textit{DRUS}} for Diffusion Reconstruction of Ultrasound images.
For comparison, we also define the \textbf{\textit{Deno}} approach, which assumes $\Bv\Hv=\Iv$ reducing the method to pure denoising.
}

\DM{The noise characteristics of the updated direct model $\Bv\nv$ may not strictly adhere to the white noise assumption, and we explored a potential workaround to address this concern in our preliminary work~\cite{zhang2023}. 
}

\DM{In our experimental setup, the diffusion model is embedded with 1000 levels of noise. While a traditional generative sampling process employing the reverse diffusion SDE (Stochastic Differential Equation) traverses all of the noise levels, our DDRM-based approach takes advantage of DDIM (Denoising Diffusion Implicit Models)~\cite{DDIM}, which exploits the generative probability flow ODE (Ordinary Differential Equation) to achieve a rapid convergence in sampling performance.}

Finally, as any other diffusion model, DRUS will produce for a given input data vector $\yv$, a non-deterministic solution of the reconstruction problem. In the following experiments we study the variability of this solution through independent realizations, or samples, of DRUS. \DM{Hereafter,} we denote with DRUSOne a mono-sample solution. To summarize the variability of multi-sample DRUS, we calculate the empirical mean and variance, respectively named, DRUSMean and DRUSVar. \DM{As we later show DRUSMean \YZ{performs similar to DRUSOne}, but 
interestingly DRUSVar can improve US image reconstruction both in terms of \YZ{contrast and Signal-to-Noise Ratio (SNR) without sacrificing resolution}. 
}

\section{Experimental Setup}
\label{sec:expeSetup}
\subsection{Training Datasets}
\DM{Several unconditional diffusion models trained on vast publicly available natural image datasets exist and have been made accessible open source}.
Nevertheless, due to the significant disparity between the data distributions of the \YZ{noise-free} natural image datasets and of the desired ultrasound reflectivity maps \YZ{embedded with the signed speckle noise}, \DM{using} a diffusion model solely trained on natural images presents an out-of-distribution challenge. Conversely, training a diffusion model exclusively on ultrasound data demands a substantial, high-quality training dataset which is challenging to acquire, especially for medical applications. As a result, we \DM{opt for} 
fine-tuning an open-source generative diffusion model~\cite{dhariwal_diffusion_2021} originally pre-trained on ImageNet~\cite{ILSVRC15} using a relatively small 
\DML{dataset of high-quality ultrasound images compounded from 65-101 plane wave transmissions}. This \DM{strategy} enables us to leverage the learned patterns from natural images \DM{while mitigating} the impact of the out-of-distribution problem without the requirement for an extensive high-quality ultrasound dataset and a long training time.

Our experiments \DM{are based on} three unconditional diffusion models with identical architectures at a resolution of $256 \times 256$, both fine-tuned from the same checkpoint\footnote{
\DM{Downloaded from:}
\underline{https://github.com/openai/guided-diffusion}}. The sole distinction between these models lies in their respective fine-tuning datasets. The model used for the simulated and experimental data (in Section \ref{subsec:valResultsPhan}) was fine-tuned on 824 high-quality unpaired \textit{in vitro} ultrasound images. For the \textit{in vivo} data (in Section \ref{subsec:valResultsVivo}, Fig.~\ref{fig: picmusImagesSummaryVivo})
, the model underwent fine-tuning on a comprehensive dataset consisting of 3551 images (824 \textit{in vitro} + 1515 \textit{in vitro} + 1212 \textit{in vivo}).  The fine-tuning dataset for the third model consists of 1012 carotid cross-sectional images, with a specific focus on the additional \textit{in vivo} data (in Section \ref{subsec:valResultsVivo}, Fig.~\ref{fig: additioanlImagesSummaryVivo}). Notably, these fine-tuning datasets were collected from a TPAC Pioneer machine. The \textit{in vitro} images were acquired using the CIRS 040GSE phantom, while the \textit{in vivo} images were obtained from the carotid artery of a volunteer. \CH{All our \textit{in vivo} acquisitions were conducted \YZ{with the consent of the volunteer} and in accordance with the ethical principles of the Declaration of Helsinki. }

\subsection{Validation Datasets}
\label{sec:valDatasets}
We conducted \DM{a quantitative and qualitative} evaluation of our approach \DM{based on} the publicly available datasets from the Plane Wave Imaging Challenge in Medical UltraSound (PICMUS)~\cite{PICMUS}, alongside 
the self-acquired \textit{in vivo} dataset.

The PICMUS datasets comprise two simulation sets, two \textit{in vitro} sets, and two \textit{in vivo} sets. All acquisitions within this challenge utilized a 128-element L11–4v linear-array transducer with a pitch of 0.30 mm, an element width of 0.27 mm, and an element height of 5 mm. A transmit pulse with a central frequency of 5.208 MHz and a bandwidth ratio (BWR) of 67\% was employed, with a sampling rate of 20.8 MHz. Data was made available in either RF (Radio Frequency) or IQ (In-Phase Quadrature) formats, with our experiments utilizing the RF format. We name each dataset as follows, according to the features of their corresponding images:
\begin{itemize}
    \item \textit{SR (Simulation Resolution)}:  image with point targets distributed both horizontally and vertically against an anechoic background, intended to evaluate spatial resolution.
    \item \textit{SC (Simulation Contrast)}: image with a total of 9 anechoic regions distributed horizontally and vertically against a speckle background, \DM{designed} to assess contrast.
    \item \textit{ER (Experimental Resolution)}: image \DM{with a series of} point targets and a hyperechoic region against a speckle background, enabling the evaluation of both spatial resolution and contrast. 
    \item \textit{EC (Experimental Contrast)}: image focuses with two anechoic regions against a speckle background, enabling the evaluation of contrast.
    \item \textit{CC (Carotid Cross)}:  carotid cross sectional view.
    \item \textit{CL (Carotid Long)}:  carotid longitudinal view.
\end{itemize}
The simulation datasets \textit{SR} and \textit{SC} were generated through the ultrasound simulation package Field\,II \cite{FieII_1,FieII_2}, the \textit{in vitro} datasets \textit{ER} and \textit{EC} were collected from a CIRS 040GSE Phantom, and the \textit{in vivo} datasets \textit{CC} and \textit{CL} were acquired from the carotid artery of a volunteer.

The additional \textit{in vivo} dataset comprises five distinct sequence sets, acquired from the carotid artery of a volunteer, utilizing a TPAC Pioneer ultrasound machine equipped with an L11-5 probe.  A transmit pulse with
a central frequency of 5.0 MHz and a BWR of 50\% was employed.

\subsection{Inverse Problem Model Parameters}
Properly constructing matrices $\Hv$ and $\Bv$ is essential for leveraging the linear inverse problem model in \eqref{Equ: model_HtH} to do image reconstruction. 
The construction of the forward matrix $\Hv$ necessitates the channel data acquisition parameters, the field of view, and the image resolution. The data acquisition parameters were set according to the PICMUS detailed in Section~\ref{sec:valDatasets}, the field of view spans from -18 mm to 18 mm in width and from 10 mm to 46 mm in depth, with the origin located at the transducer's center. The image resolution was fixed at $256 \times 256$.
The construction of the beamforming matrix $\Bv$ relies on the same parameters 
\DM{as $\Hv$, 
\YZ{ and the receive apodization weights defined by a window of Tukey0.25 and an \textit{f-number} of 1.4.}}
Since the data acquisition parameters of the PICMUS and our \textit{in vivo} validation datasets are similar, we used the same model matrix to conduct all of the experiments.
 
\subsection{Evaluation Metrics}
\DM{Following the PICMUS standardized evaluation~\cite{PICMUS},}
we assess the quality of the reconstructed ultrasound images using the following 
metrics:

1) Axial and Lateral Resolution: 
Measured as the -6dB Full Width at Half Maximum (FWHM) on the bright scatterers within the \textit{SR} and the \textit{ER} images. A smaller FWHM value indicates higher resolution.

2) Contrast: We rely on both the Contrast to Noise Ratio (CNR) and the generalized Contrast to Noise Ratio (gCNR)~\cite{gCNR} metrics. 
The CNR is calculated as:
$$\mathrm{CNR}=10 \log_{10}\bigg(\frac
{\left|\mu_{\text{in}}-\mu_{\text{out}}\right|^2}{\left(\sigma_{\text {in }}^2+\sigma_{\text {out}}^2\right) / 2}\bigg),
$$
where the subscripts `in' and `out' indicate inside or outside the target regions, $\mu$ and $\sigma$ denote the mean and the standard deviation respectively. The gCNR that has been shown to be robust against dynamic range alterations is calculated as:
$$
\mathrm{gCNR}=1-\int_{-\infty}^{\infty} \min \left\{g_{\text {in}}(v), g_{\text {out}}(v)\right\} dv,
$$
where $v$ denotes the pixel values, and $g$ refers to the histogram of pixels in each region.
Contrast is measured on the hyperechoic region within the \textit{ER} image and on the anechoic regions within the \textit{SC} and the \textit{EC} images. Higher CNR values and a gCNR closer to 1 indicate superior contrast.

3) Background quality: This metric is measured as the SNR $\mu_\text{ROI} / \sigma_\text{ROI}$ where the subscript `ROI' refers to the region of interest, and the p-value of the Kolmogorov–Smirnov (KS) test under a Rayleigh distribution hypothesis. The background quality is evaluated on both \textit{SC} and \textit{EC} images.
An SNR value of 
$\sim$1.91 and a p-value greater than 0.05 are indicative of well-preserved ultrasound speckle texture, a characteristic often desired in medical applications. However, considering the forward-looking perspective presented in Task 1b of the Challenge on Ultrasound Beamforming with Deep Learning (CUBDL)\cite{CUBDL}, achieving the highest possible SNR, irrespective of speckle preservation, also holds significance.


\section{Experimental Validation and Results}
\label{sec:valResults}
\subsection{Phantom based performance assesment}
\label{subsec:valResultsPhan}
\subsubsection{Sensitivity of Deno and DRUS to the iteration step count for single-sample reconstructions}
\begin{figure*}[ht]
\centering
    \includegraphics[width=0.49\textwidth]{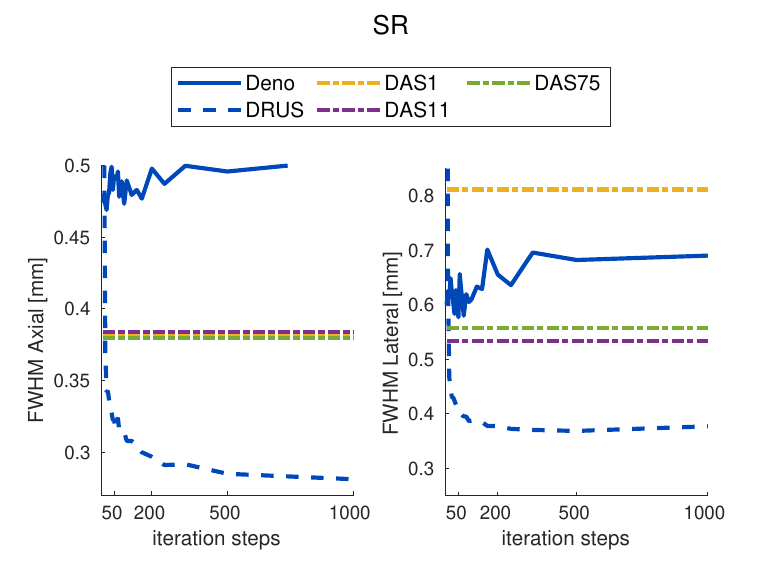}
    \includegraphics[width=0.49\textwidth]{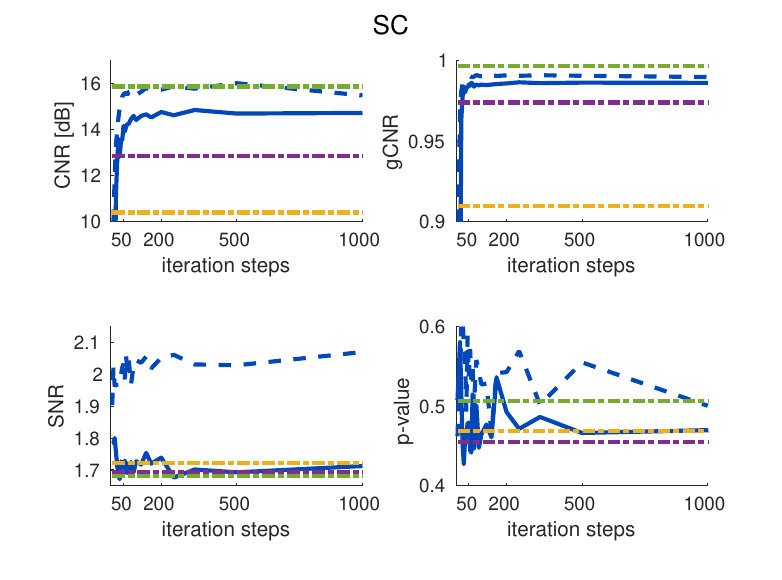}
    \includegraphics[width=0.49\textwidth]{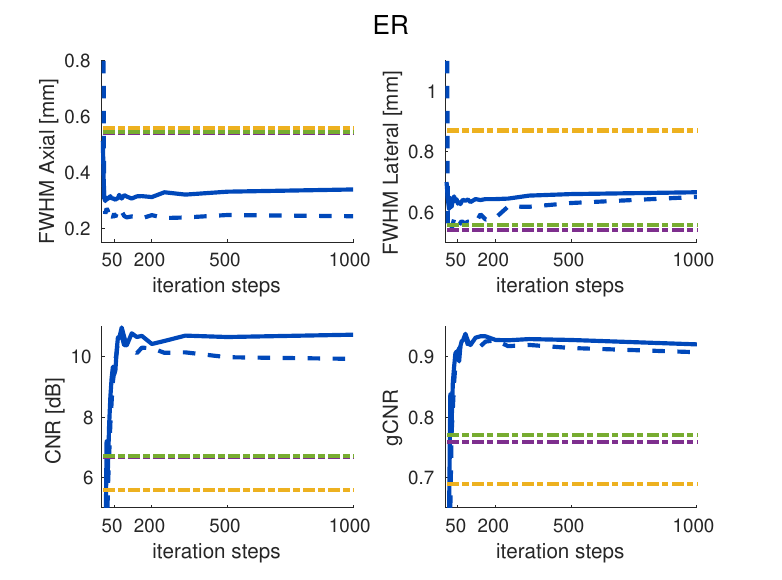}
    \includegraphics[width=0.49\textwidth]{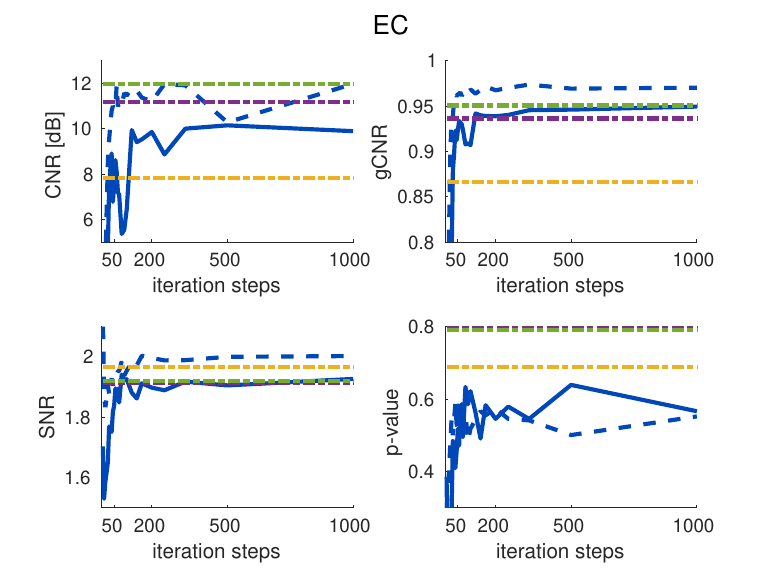}
\caption{Sensitivity of Deno and DRUS to the iteration step count in the sampling process, evaluated on PICMUS, and compared to DAS. 
} 
\label{fig: itSensitivity}
\end{figure*}

In the context of natural image restoration tasks, 
DDRM has exhibited satisfactory experimental outcomes with merely 20 iteration steps. However, our empirical investigations prove that the iteration step count of 20 is frequently insufficient for DRUS to attain quality convergence. To assess the impact of \textit{it} on our method, we conducted an analysis on the reconstructed PICMUS images at different values of this parameter. The averaged metrics are presented in Fig.~\ref{fig: itSensitivity}. 

The curves of DRUS and Deno in Fig.~\ref{fig: itSensitivity} show a rapid convergence inherent \DM{to} DDRM.  In the context of restoring a medical ultrasound image, the elbow regions of these curves confirm the adequacy of an \textit{it} ranging from 50 to 100 while corroborating the redundancy of using more than 200 \textit{it} steps. 

The 
advantage of DRUS over Deno is also illustrated in Fig.~\ref{fig: itSensitivity}. DRUS achieves competitive or superior scores in comparison to the golden standard DAS75 (DAS with 75 Plane Wave (PW) transmissions)\DM{~\cite{PICMUS,CUBDL}} across all metrics, except for the p-value of \textit{EC}. Although Deno exhibits slightly better performance than DRUS in terms of \textit{ER}'s contrast and \textit{EC}'s p-value, the 
performance gap between Deno and DRUS is evident, particularly in the case of \textit{SR}, where Deno falls short even when compared to DAS1 (DAS with 1 PW).

\subsubsection{Statistical behavior of the reconstruction}\label{sec: histograms}

\begin{figure*}[ht]
\centering
    \includegraphics[width=0.59\textwidth]{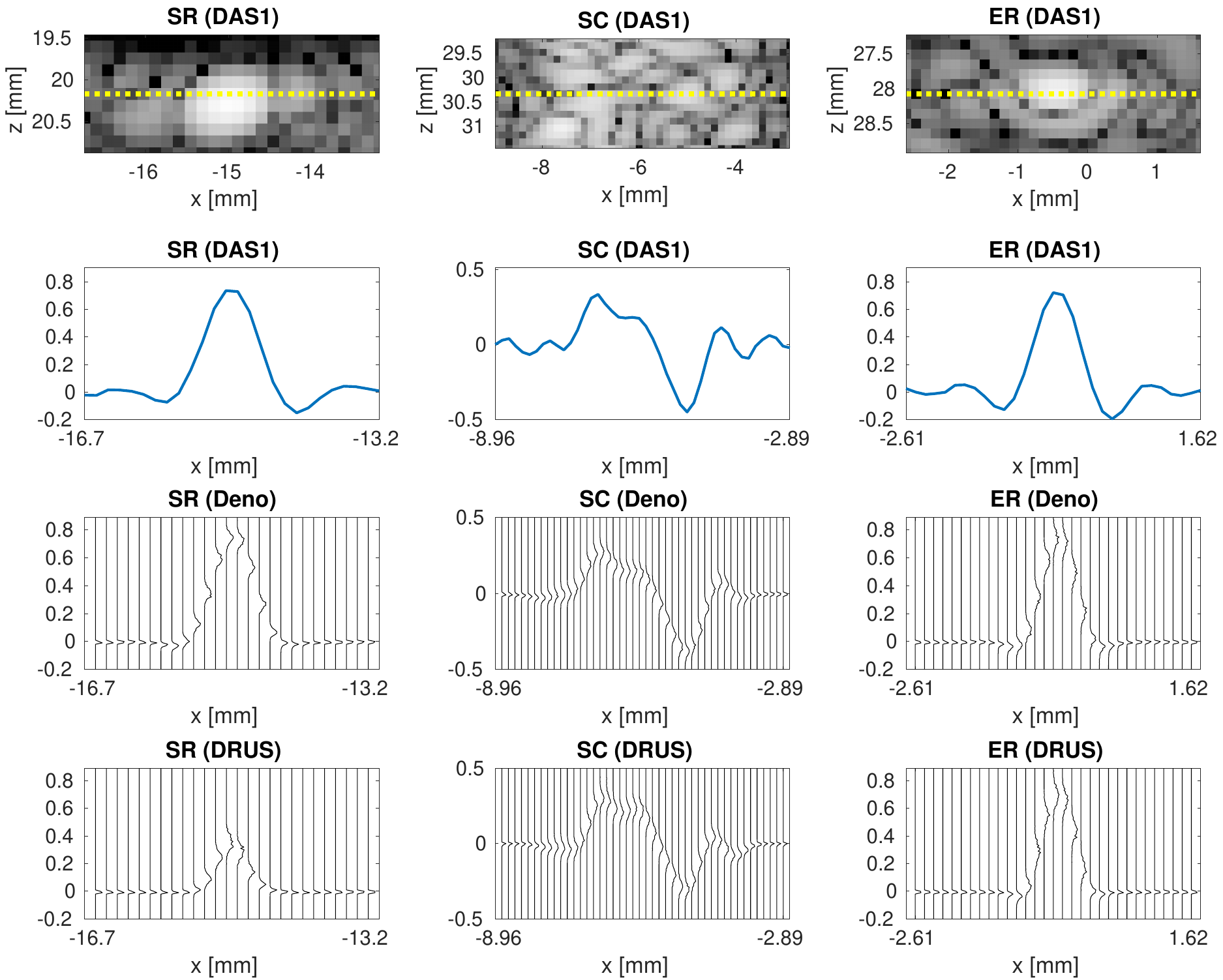}
    \includegraphics[width=0.395\textwidth]{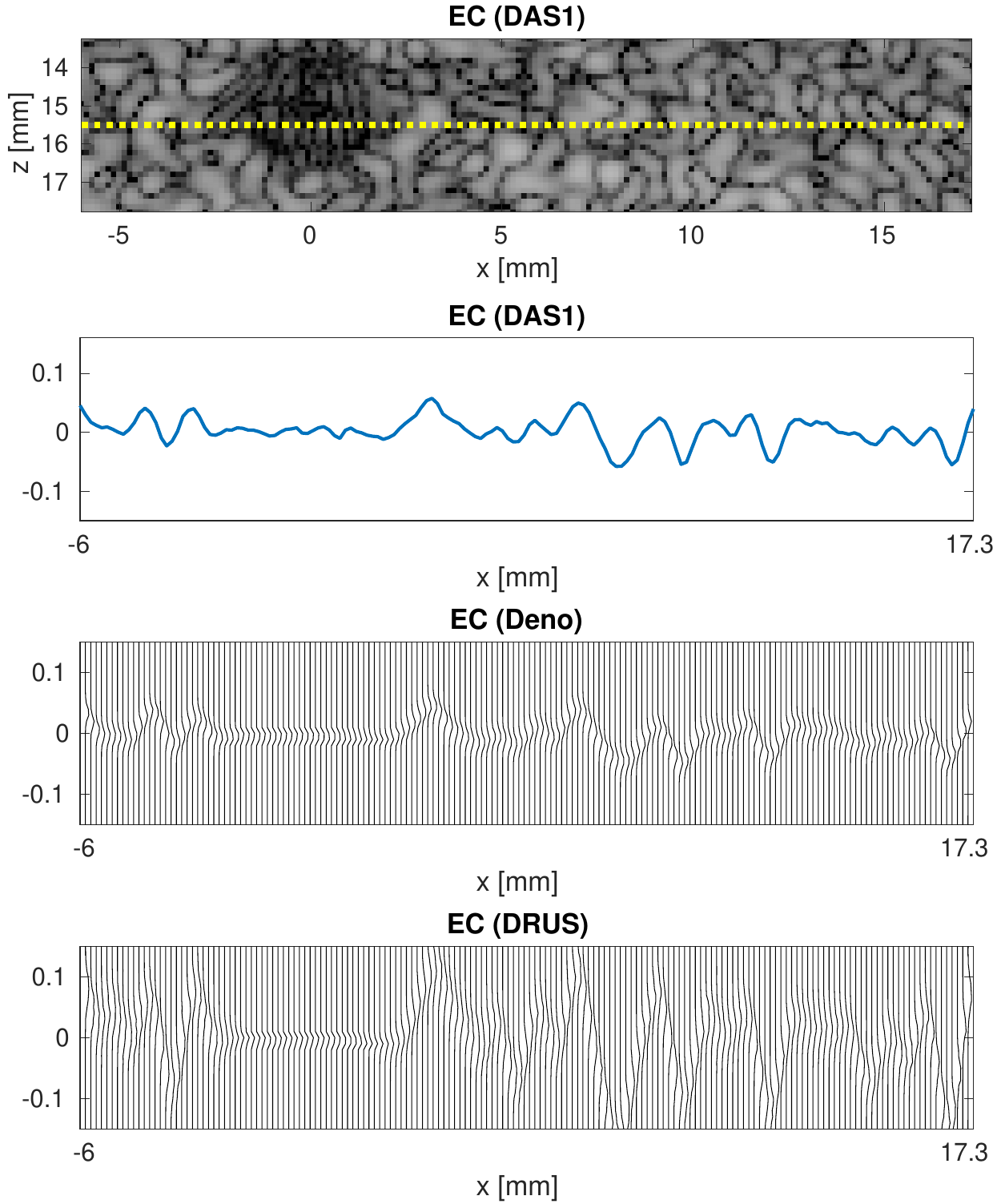}
\caption{Statistical behaviour of Deno and DRUS compared with the DAS1 image on PICMUS.} 
\label{fig: histograms}
\end{figure*}

To have a better knowledge of the uncertainty of our method, we conducted an analysis of the statistical properties of our approach with multiple samplings. In Fig.~\ref{fig: histograms}, we display local regions of the DAS1 
beamformed PICMUS images \DML{(top-row)}, plot 
\DML{as a function (second-row) the intensity values of the line profile in yellow,}
and present the histograms of multiple samplings using Deno and DRUS.

The histograms in Fig.~\ref{fig: histograms} reveal a  generative uncertainty with a normal distribution pattern. Furthermore, the consistency between the histograms' mean values and the DAS1 pixel value curves serves as evidence of Deno and DRUS's ability to preserve background speckles.

Additionally, the variances of the histograms exhibit \DM{interesting} patterns: within the hypoechoic regions, both Deno and DRUS yield a stable 
%
\DM{zero mean and a variance also close to zero;}
within the hyperechoic regions, such as in \textit{SR} and \textit{ER}, the histogram's mean rapidly goes up at the scatterer, accompanied by a significant increase in variance. 
\DML{For the more complex background speckle of}
 \textit{SC} and \textit{EC}, the variance remains relatively constant even as the mean undergoes gradual fluctuations. 

\subsubsection{Imaging with multiple samples}\label{sec: multiSamples}
\begin{figure*}[ht]
\centering
    \includegraphics[width=0.49\textwidth]{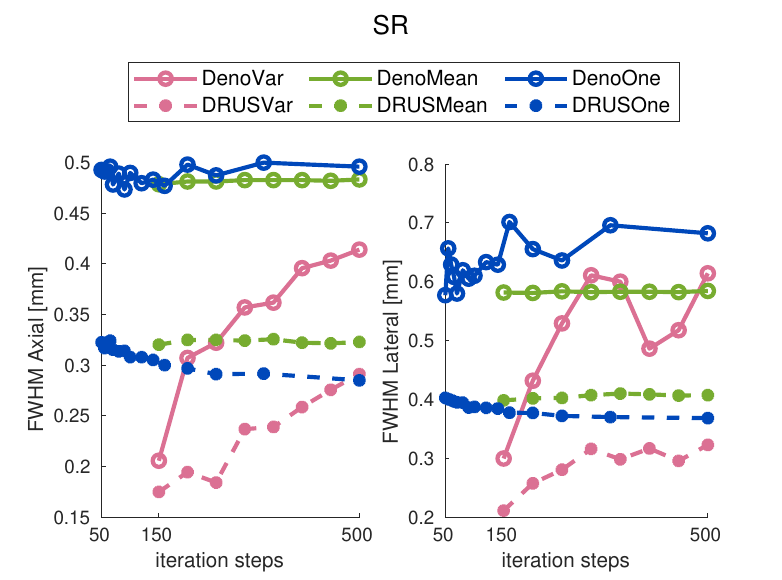}
    \includegraphics[width=0.49\textwidth]{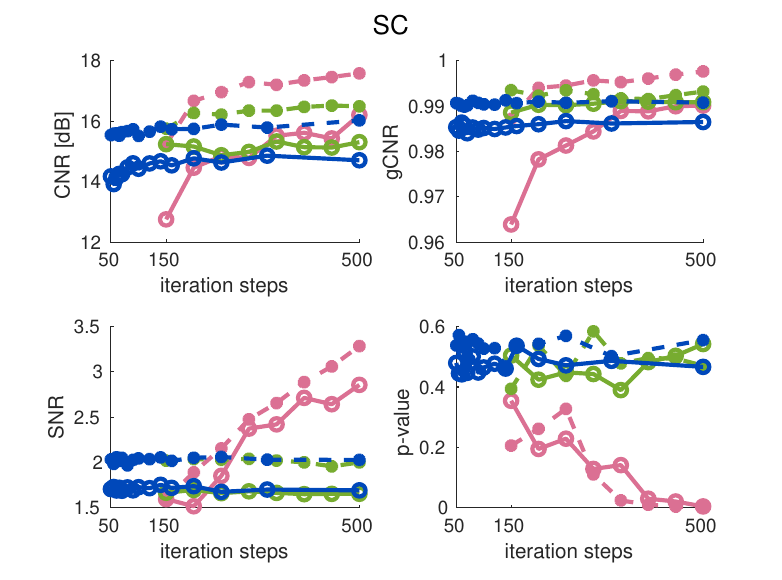}
    \includegraphics[width=0.49\textwidth]{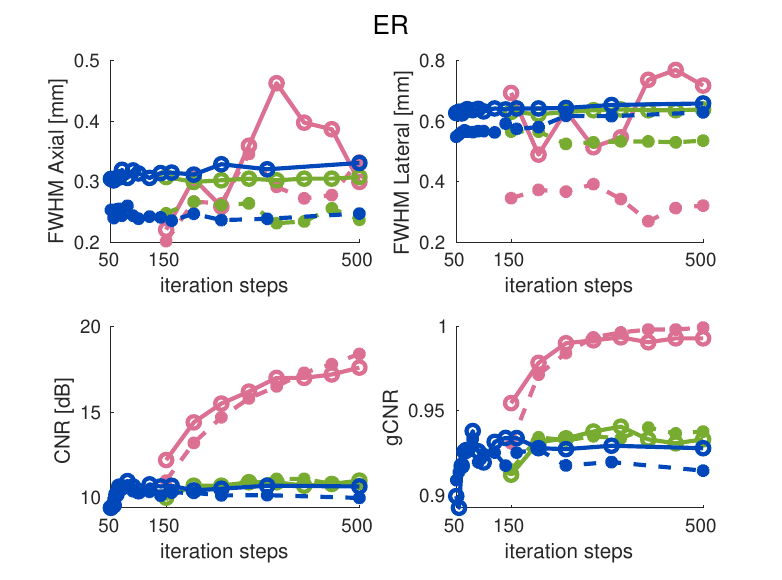}
    \includegraphics[width=0.49\textwidth]{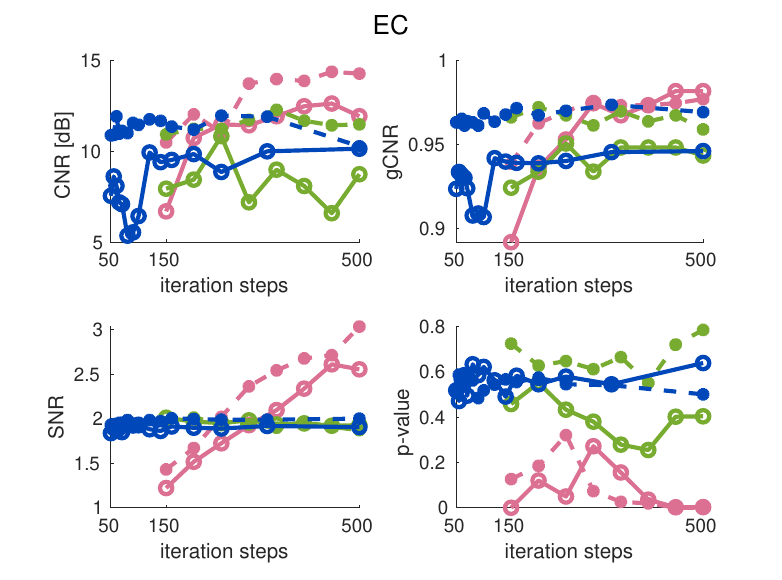}
\caption{\DM{
 Quantitative comparison of single- and multiple-sample images on PICMUS. In the multiple-sample case, each sample relies on 50 \textit{it}.}
}

\label{fig: Mul_vs_One50it}
\end{figure*}

\begin{figure*}[ht]
\centering
\includegraphics[width=\textwidth]{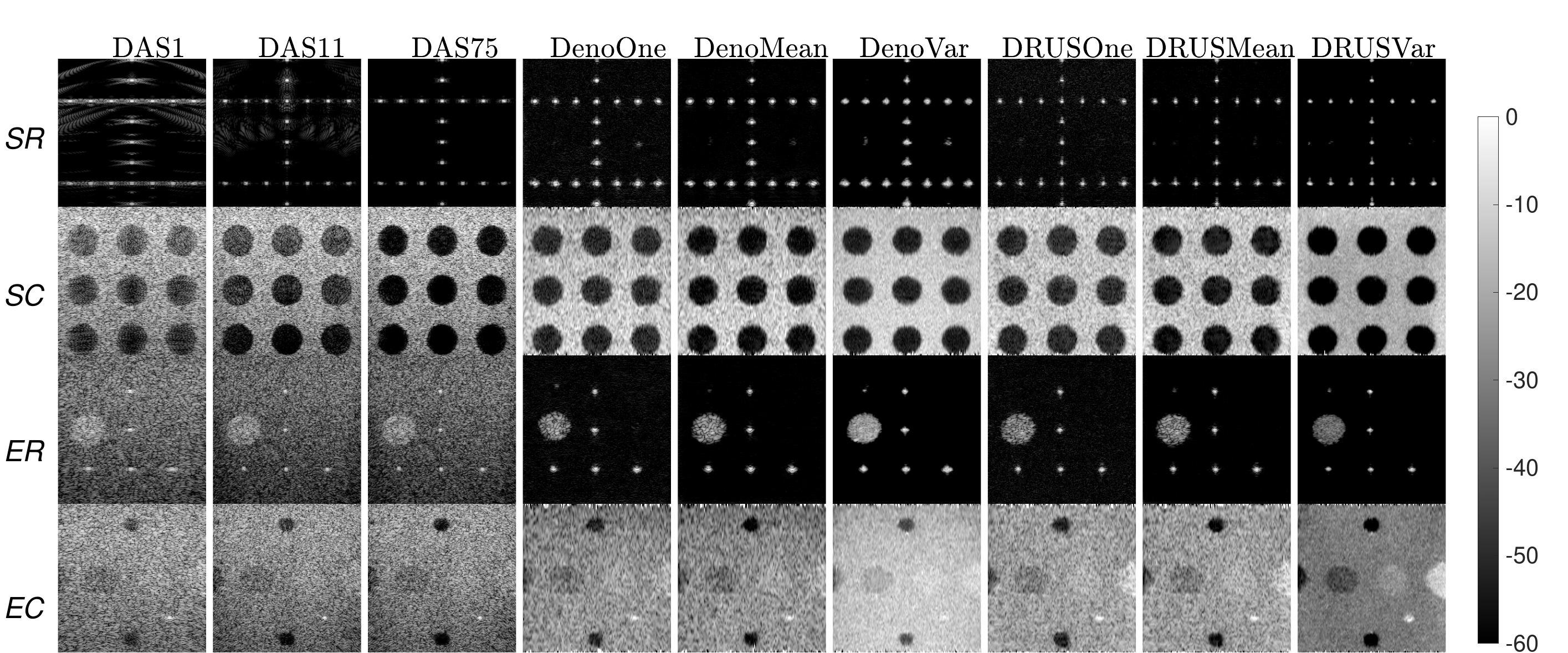}
\caption{Qualitative comparison of the reconstructed phantom-based PICMUS images. All images are in decibels with a dynamic range [-60,0].} 
\label{fig: picmusImagesSummary}
\end{figure*}

\DM{Here, we further explore the
positive correlation 
observed in Section}~\ref{sec: histograms}
between variance and the rate of pixel value variations, a relationship we exploit in the construction \DML{of DRUSVar images}.\YZ{The variance images, the mean images, and the single samples of DRUS and Deno} are compared quantitatively and qualitatively in Fig.~\ref{fig: Mul_vs_One50it} and in Fig.~\ref{fig: picmusImagesSummary} respectively.

 \DM{In Fig.~\ref{fig: Mul_vs_One50it}, we compare the single and multiple-sample approaches on the PICMUS dataset for an increasing number of iterations. For a fair comparison in terms of computational resources, we let single-sample models vary between 50 and 500 iterations. Then,  for multiple-sample approaches, we fix each sampled reconstruction to 50 iterations and add up the total number of iterations required to go from 1 to 10 reconstructed images} \DML{(used to compute the mean and variance).}

Fig.~\ref{fig: Mul_vs_One50it} first underscores the superiority of DRUS over Deno across various performance indicators.
Fig.~\ref{fig: Mul_vs_One50it} also shows that the metrics of the mean images \YZ{(DRUSMean, DenoMean)} are 
\DML{relatively}
stable to the sampling count and are close to the metrics of the single samples \YZ{(DRUSOne, DenoOne)}. However, the \YZ{metrics of the variance images (DRUSVar, DenoVar) exhibit significant alterations as the sampling count increases from 1 to 10, }
particularly about the improvement of contrast and SNR, and the reduction of speckle pattern. For \textit{ER}, it is observed that DRUSVar starts to surpass DRUSMean or DRUSOne with only three rounds of sampling. As the number of samples exceeds three, the contrast of DRUSVar continues to rise without compromising the resolution, while the scores of DRUSMean or DRUSOne remain stable. For \textit{SC} and \textit{EC}, DRUSVar surpasses the peak performance of DRUSMean or DRUSOne in terms of CNR, gCNR, and SNR with only five to six rounds of sampling. Additionally, DRUSVar maintains a consistent advantage in resolution metrics for \textit{SR}.

Fig.~\ref{fig: picmusImagesSummary} 
is congruent
with the metrics 
in Fig.~\ref{fig: Mul_vs_One50it}, affirming the feasibility of DRUSVar \DM{as a reconstruction method}
that
effectively enhances contrast, improves SNR, and attenuates background speckles without compromising resolution.

\subsubsection{Comparison to state-of-the-art}
In addition to comparing our method, DRUS, with Deno and DAS (1PW, 11PWs, and 75 PWs), \DM{we report} a quantitative comparison against six other techniques in Table \ref{tab: summary}, including the EMV~\cite{asl_eigenspace-based_2010},
the PCF~\cite{PCF}, a model-based approach 
with regularization by denoising (RED)~\cite{RED_USIPB}, and three learning-based approaches MNV2~\cite{MNV2}, ABLE~\cite{luijten_adaptive_2020} and DNN-$\lambda^*$~\cite{ZHANG2021}. 

\begin{table*}[ht]
\caption{Quantitative comparison to the state-of-the-art on the PICMUS phantom-based datasets. A and L denote axial and lateral directions respectively. All of the samples of Deno and DRUS underwent 50 iteration steps and the variance images are constructed with 10 samples. The best values are bolded.}
\label{tab: summary}
\resizebox{\textwidth}{!}{%
\begin{tabular}{ccc|ccc|cc|cc|cccccc}
\hline
 &  &  & \multicolumn{3}{c|}{DAS} & \multicolumn{2}{c|}{One Sampling} & \multicolumn{2}{c|}{Variance Imaging} & \multirow{2}{*}{EMV~\cite{asl_eigenspace-based_2010}} & \multirow{2}{*}{PCF~\cite{PCF}} & \multirow{2}{*}{RED~\cite{RED_USIPB}} & \multirow{2}{*}{MNV2~\cite{MNV2}} & \multirow{2}{*}{ABLE~\cite{luijten_adaptive_2020}} & \multirow{2}{*}{DNN-$\lambda^*$~\cite{ZHANG2021}} \\
 &  &  & 1 & 11 & 75 & Deno & DRUS & Deno & DRUS &  &  &  &  &  &  \\ \hline
\multicolumn{1}{c|}{\multirow{2}{*}{\textit{SR}}} & \multirow{2}{*}{\begin{tabular}[c]{@{}c@{}}FWHM\\ {[}mm{]}\end{tabular}} & A$\downarrow$ & 0.38 & 0.38 & 0.38 & 0.49±0.02 & 0.32±0.01 & 0.41 & 0.29 & 0.40 & 0.30 & 0.37 & 0.42 & \textbf{0.22} & 0.28 \\
\multicolumn{1}{c|}{} &  & L$\downarrow$ & 0.81 & 0.53 & 0.56 & 0.58±0.01 & 0.40±0.02 & 0.61 & 0.32 & \textbf{0.10} & 0.38 & 0.46 & 0.27 & 0.70 & 0.32 \\ \hline
\multicolumn{1}{c|}{\multirow{4}{*}{\textit{SC}}} & \multicolumn{2}{c|}{CNR{[}dB{]}$\uparrow$} & 10.41 & 12.86 & 15.89 & 14.18±0.29 & 15.55±0.21 & 16.21 & \textbf{17.59} & 11.21 & 0.46 & 15.48 & 10.48 & 11.91 & 10.85 \\
\multicolumn{1}{c|}{} & \multicolumn{2}{c|}{gCNR$\uparrow$} & 0.91 & 0.97 & \textbf{1.00} & 0.99±0.00 & 0.99±0.00 & 0.99 & \textbf{1.00} & 0.93 & 0.41 & 0.94 & 0.89 & / & / \\
\multicolumn{1}{c|}{} & \multicolumn{2}{c|}{SNR$\uparrow$} & 1.72 & 1.69 & 1.68 & 1.71±0.08 & 2.03±0.07 & 2.86 & \textbf{3.28} & / & / & / & / & / & / \\
\multicolumn{1}{c|}{} & \multicolumn{2}{c|}{KS p-value} & 0.47/\checkmark & 0.46/\checkmark & 0.51/\checkmark & 0.48±0.15/\checkmark & 0.54±0.10/\checkmark & 0.00/\xmark & 0.00/\xmark & \checkmark & \xmark & \checkmark & \checkmark & / & / \\ \hline
\multicolumn{1}{c|}{\multirow{4}{*}{\textit{ER}}} & \multirow{2}{*}{\begin{tabular}[c]{@{}c@{}}FWHM\\ {[}mm{]}\end{tabular}} & A$\downarrow$ & 0.56 & 0.54 & 0.54 & 0.30±0.02 & \textbf{0.25±0.02} & 0.30 & 0.34 & 0.59 & 5.64 & 0.48 & 0.53 & / & 0.52 \\
\multicolumn{1}{c|}{} &  & L$\downarrow$ & 0.87 & 0.54 & 0.56 & 0.63±0.02 & 0.55±0.03 & 0.72 & \textbf{0.32} & 0.42 & 0.76 & 0.76 & 0.77 & / & 0.52 \\
\multicolumn{1}{c|}{} & \multicolumn{2}{c|}{CNR{[}dB{]}$\uparrow$} & 5.60 & 6.70 & 6.70 & 9.41±0.98 & 9.50±0.32 & 17.60 & \textbf{18.40} & / & / & / & / & / & / \\
\multicolumn{1}{c|}{} & \multicolumn{2}{c|}{gCNR$\uparrow$} & 0.69 & 0.76 & 0.77 & 0.90±0.04 & 0.91±0.01 & 0.99 & \textbf{1.00} & / & / & / & / & / & / \\ \hline
\multicolumn{1}{c|}{\multirow{4}{*}{\textit{EC}}} & \multicolumn{2}{c|}{CNR{[}dB{]}$\uparrow$} & 7.85 & 11.20 & 12.00 & 7.56±2.23 & 10.90±1.24 & 11.95 & 14.30 & 8.10 & 3.20 & \textbf{14.70} & 7.80 & / & 11.6 \\
\multicolumn{1}{c|}{} & \multicolumn{2}{c|}{gCNR$\uparrow$} & 0.87 & 0.94 & 0.95 & 0.92±0.02 & 0.96±0.01 & \textbf{0.98} & \textbf{0.98} & 0.83 & 0.68 & \textbf{0.98} & 0.83 & / & / \\
\multicolumn{1}{c|}{} & \multicolumn{2}{c|}{SNR$\uparrow$} & 1.97 & 1.91 & 1.92 & 1.84±0.06 & 1.93±0.08 & 2.55 & \textbf{3.03} & / & / & / & / & / & / \\
\multicolumn{1}{c|}{} & \multicolumn{2}{c|}{KS p-value} & 0.69/\checkmark & 0.80/\checkmark & 0.79/\checkmark & 0.52±0.14/\checkmark & 0.52±0.16/\checkmark & 0.00/\xmark & 0.00/\xmark & \checkmark & \xmark & \checkmark & \checkmark & / & / \\ \hline
\end{tabular}%
}
\end{table*}
As seen in Table.~\ref{tab: summary}, in terms of resolution and contrast, our method, DRUS, is overall significantly better than DAS1 and can compete with DAS75 and the state-of-the-art approaches. When background speckle preservation is not a concern, \YZ{DRUSVar} 
is a good choice to yield images characterized by significantly elevated SNR and enhanced contrast.

\subsection{Performance on \textit{in vivo} data}
\label{subsec:valResultsVivo}
\begin{figure*}[ht]
\centering
\includegraphics[width=0.95\textwidth]{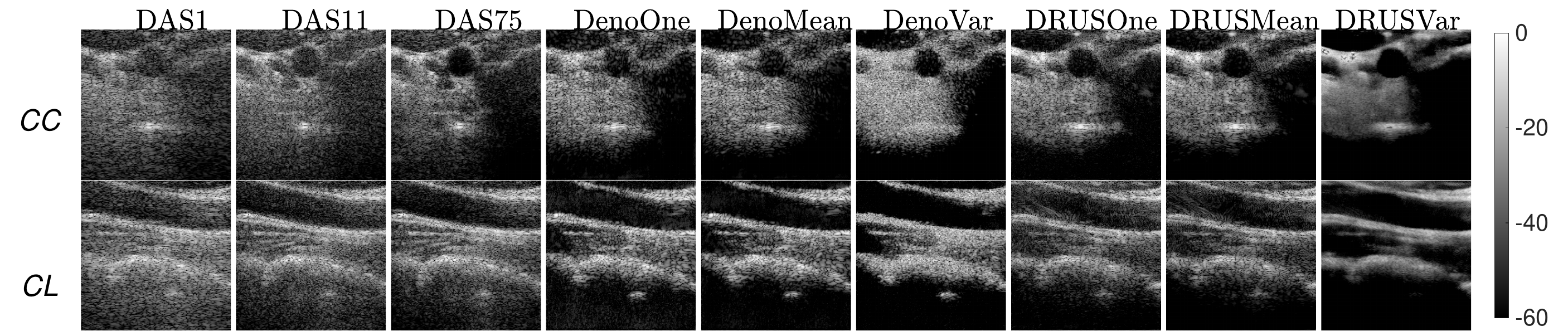}
\caption{
Comparison of reconstructed images
on the PICMUS 
\textit{in vivo} datasets. 
All images are in decibels with a dynamic range [-60,0].
} 
\label{fig: picmusImagesSummaryVivo}
\end{figure*}

\begin{figure*}[ht]
\centering
\includegraphics[width=0.85\textwidth]{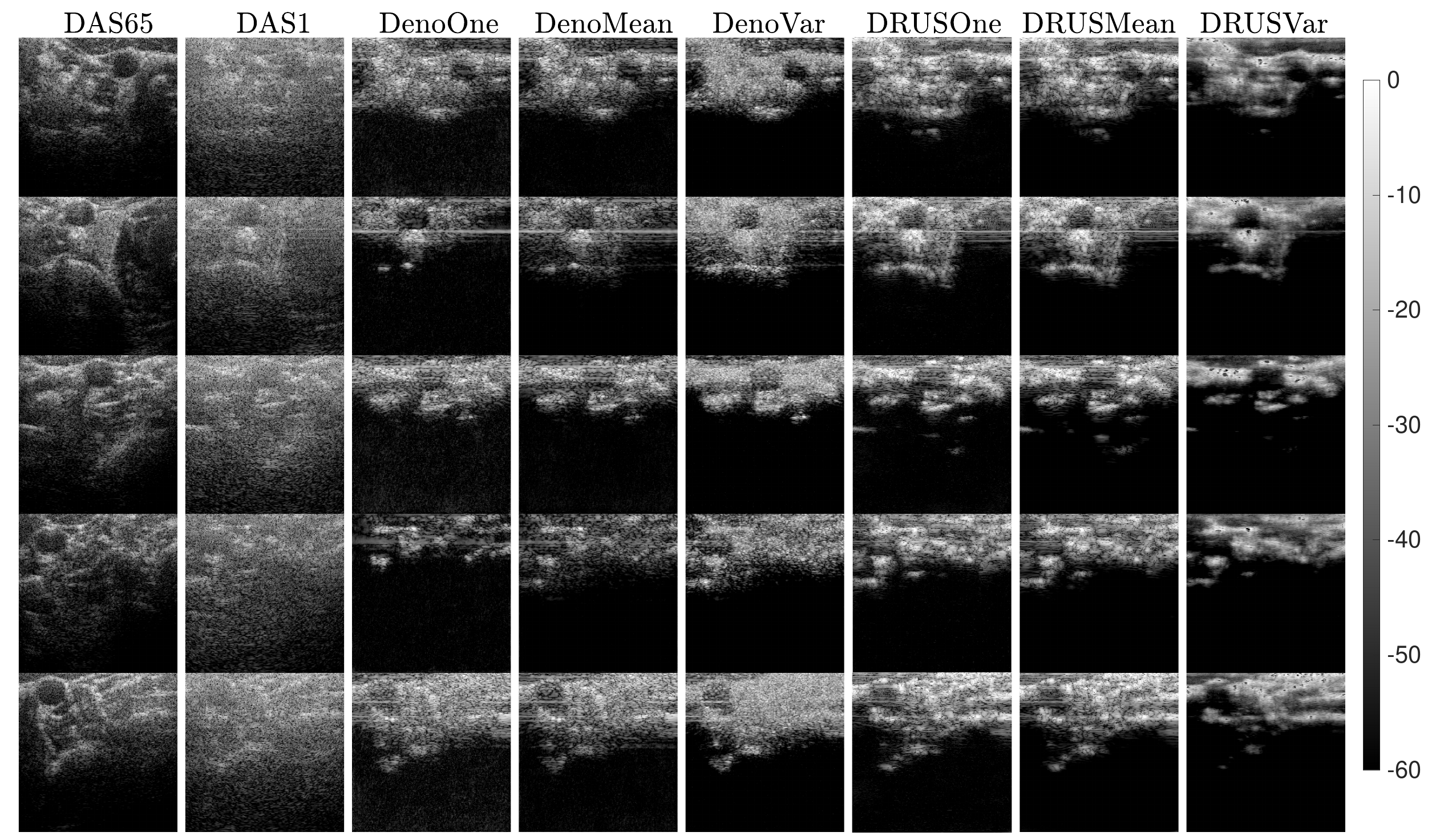}
\caption{\DM{Comparison of reconstructed images from} the additional \textit{in vivo} datasets. 
All images are in decibels with a dynamic range [-60,0].} 
\label{fig: additioanlImagesSummaryVivo}
\end{figure*}

\DML{Next, we qualitatively} compare DRUS with Deno and DAS using \textit{in vivo} datasets, as presented in Fig.~\ref{fig: picmusImagesSummaryVivo} and Fig.~\ref{fig: additioanlImagesSummaryVivo}. Deno and DRUS underwent 4 and 20 rounds of sampling respectively to construct the images of the mean and the variance. We empirically observed that increasing the sampling count for DenoVar led to a degradation of image details.

As depicted in Fig.~\ref{fig: picmusImagesSummaryVivo} and Fig.~\ref{fig: additioanlImagesSummaryVivo}, both single-sample and multi-sample imaging approaches of Deno and DRUS effectively reduce additive noise compared to DAS1. Notably, DRUS outperforms Deno, particularly on the PICMUS CC dataset. When considering the multi-sample variance-based imaging approach, DenoVar exhibits a significant loss of image details even with only four samples, while DRUSVar retains image details and simultaneously achieves background speckle removal and additive noise reduction.

\section{Discussion}
\label{sec:discussion}
In the domain of medical ultrasound image reconstruction, some studies \cite{RED_USIPB,ZHANG2021} 
\DM{focus}
on restoring the reflectivity map, while others \cite{Hyun19,usDespeckle2022Lee} 
\DM{opt for}
reconstructing the echogenicity map. 
A frequently employed model \cite{speckleModel2007Ng} depicting the association between the reflectivity map, denoted as $\ov$, and the echogenicity map, denoted as $\pv$, is: 
\begin{align}
    \ov = \Dv \pv,
\label{Equ: o=Dp}
\end{align}
where \YZ{$\Dv \in \R^{N\times N}$ is a diagonal matrix,}
\DML{with diagonal elements sampled from}
the standard normal distribution, representing the multiplicative noise responsible for the speckle. 

Based on the DRUS results in Section \ref{sec:valResults}, the reconstructed images formed by DRUSMean bear a closer resemblance to a reflectivity map while DRUSVar appears as the echogenicity map. 
\DML{To futher investigate this behaviour,}
we conducted an analytical assessment on a publicly accessible fetal image sourced from the Field II website \cite{FieII_1,FieII_2}. In this analysis, we designated this image as the echogenicity map and synthesized a corresponding reflectivity map by applying \eqref{Equ: o=Dp}.

\begin{figure}[t]
\centering
(a){\includegraphics[width=0.7\linewidth]{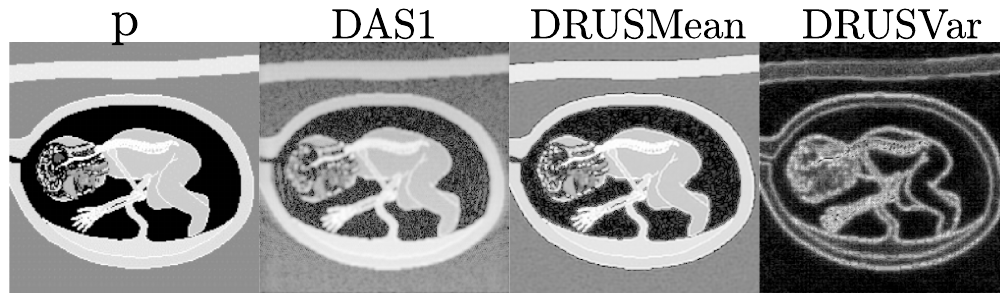}}
\\
(b){\includegraphics[width=0.7\linewidth]{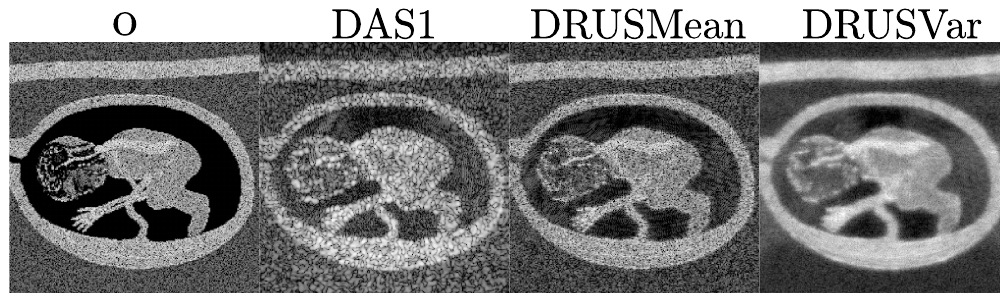}}
\caption{Qualitative comparison of image reconstructions using DAS1, DRUSMean, and DRUSVar on a simulated fetus dataset under two scenarios: (a) Echogenicity Map and (b) Reflectivity Map as the ground truths. All images are in decibels with a dynamic range [-60,0].}
\label{fig:fetus} 
\end{figure}

In a theoretical scenario where the ground truth is devoid of multiplicative noise, the forward model can be expressed as $\yv = \Hv \pv + \nv$. As depicted in Fig.~\ref{fig:fetus} (a), when comparing 
DAS1 and DRUSMean with DRUSVar, it is evident that the former two methods can reconstruct a more detailed representation of the true underlying structure $\pv$, whereas DRUSVar primarily delineates the boundaries.
This 
\DML{behaviour} 
aligns with the 
generative uncertainty observed 
on natural images 
\cite{horwitz2022conffusion}.

In the practical context \CH{of ultrasound, based on} the forward model $\yv = \Hv \ov + \nv$, a qualitative comparison 
shows that the DRUSMean image closely resembles the reflectivity map (Fig.~\ref{fig:fetus} (b)), while the DRUSVar image exhibits a stronger correspondence with the echogenicity map \YZ{(Fig.~\ref{fig:fetus} (a))}. \CH{Nevertheless}, the DRUSVar image struggles to capture low $\ov$ values in the in light gray region.


 To characterize the above statistical properties of DRUS, we introduce 
\DM{the following}
empirical model: 
\DM{}
\begin{align}
\hat\ov_r = \ov + \pv^\beta \Gv_r,
\end{align}
Here, $\hat\ov_r$ represents the $r^{th}$ reconstructed sample, $\Gv_r$ follows a standard normal distribution to account for generative uncertainty, and $\beta$ denotes an empirical parameter. It can be easily checked that $\ED[\hat\ov_r] = \ov$ and $\mathrm{Var}[\hat\ov_r] = \ov^{2\beta}$. Consequently, the echogenicity map can be approximated as $\left(\mathrm{Var}[\hat\ov_r]\right)^{1/(2\beta)}$ under this model. $\beta=0.5$ empirically yields the most favorable results with the conventional dynamic range of [-60, 0], which corresponds to DRUSVar. \CH{This simple model describes well the statistical behavior of our implementation of DDRM in the context of ultrasound image reconstruction.}

\begin{figure}
    \centering
 \includegraphics[width=0.55\columnwidth]{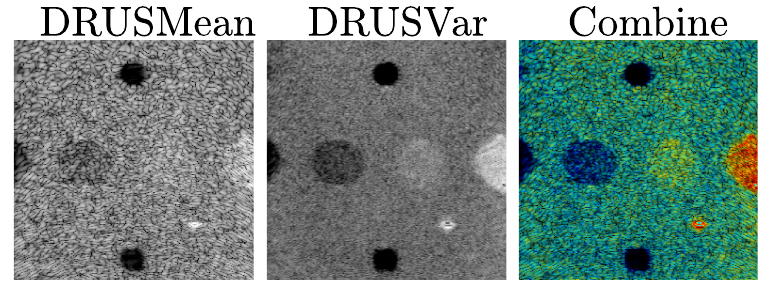}
    \caption{Example  image combining intensity-coded DRUSMean and color-coded DRUSVar images on the \textit{EC} phantom. 
    \textit{jet}  colormap.
    }   \label{fig:EC_combine}
\end{figure}

Considering our results, DRUSVar exibhits good performance both in terms of resolution and contrast, but the absence of speckle texture may be a drawback in biomedical imaging. However, by calculating DRUSVar, we have access to DRUSMean, which embodies the speckle information. In Fig.~\ref{fig:EC_combine} we propose a 
\DML{joint visualization using
DRUSMean and DRUSVar images to respectively code the luminance and chrominance of a single image. Whereas it requires a colored display, this simple representation has the advantage of a visually better contrast and resolution while keeping the speckle structure.}



\subsection{Computational time}
An NVIDIA Quadro RTX 3000 GPU requires
approximately 3-4 minutes to complete 50\textit{it} for the generation of a single sample of DRUS. An NVIDIA A100 GPU significantly accelerates this process, reducing the time to approximately 40s per sample. Importantly, when computational resources are readily available, it is feasible to 
\DM{generate multiple samples in parallel
and produce a variance image within the same computational timeframe as a single sample.}

Compared to other techniques, our method, DRUS, is slower than Deno, DAS1, PCF~\cite{PCF}, MNV2~\cite{MNV2},  ABLE~\cite{luijten_adaptive_2020} and DNN-$\lambda^*$~\cite{ZHANG2021}, but faster than EMV~\cite{asl_eigenspace-based_2010} and RED~\cite{RED_USIPB}, which need 8 and 20 minutes, respectively. RED~\cite{RED_USIPB} is slow because each iteration contains an inner iteration while EMV~\cite{asl_eigenspace-based_2010} spends time on covariance matrix evaluation and decomposition. A single sample of DRUS requires multiple multiplication operations with the singular vector matrix, which currently hinders real-time imaging. Accelerating this process is one of our key focuses for future work.

\section{Conclusion}
\label{sec:conclusion}


\DML{We propose a model-based ultrasound image reconstruction method, DRUS, that relies on a diffusion model as prior.
Building on DDRM, we incorporated ultrasound physics into the score sampling approach through the spectral decomposition of an adapted linear direct model and the unsupervised fine-tuning of the prior diffusion model.}
We conducted comprehensive evaluations on
simulated, in vitro, and in vivo datasets showing
the superior or comparable performance 
of DRUS over diffusion denoising and other state-of-the art algorithms.
Furthermore, we conducted an in-depth analysis of the statistical behaviour of DRUS, revealing, for the first time, that computing the variance of multiple samples of DRUS can achieve despeckling and result in an image with higher SNR and contrast.

\bibliographystyle{IEEEtran} 
\bibliography{IEEEabrv,references} 

\begin{thebibliography}{10}
\providecommand{\url}[1]{#1}
\csname url@samestyle\endcsname
\providecommand{\newblock}{\relax}
\providecommand{\bibinfo}[2]{#2}
\providecommand{\BIBentrySTDinterwordspacing}{\spaceskip=0pt\relax}
\providecommand{\BIBentryALTinterwordstretchfactor}{4}
\providecommand{\BIBentryALTinterwordspacing}{\spaceskip=\fontdimen2\font plus
\BIBentryALTinterwordstretchfactor\fontdimen3\font minus \fontdimen4\font\relax}
\providecommand{\BIBforeignlanguage}[2]{{%
\expandafter\ifx\csname l@#1\endcsname\relax
\typeout{** WARNING: IEEEtran.bst: No hyphenation pattern has been}%
\typeout{** loaded for the language `#1'. Using the pattern for}%
\typeout{** the default language instead.}%
\else
\language=\csname l@#1\endcsname
\fi
#2}}
\providecommand{\BIBdecl}{\relax}
\BIBdecl

\bibitem{DAS}
V.~Perrot, M.~Polichetti, F.~Varray, and D.~Garcia, ``So you think you can {DAS}?'' \emph{Ultrasonics}, vol. 111, p. 106309, Mar. 2021.

\bibitem{asl_eigenspace-based_2010}
B.~M. Asl and A.~Mahloojifar, ``Eigenspace-based minimum variance beamforming applied to medical ultrasound imaging,'' \emph{IEEE Trans. Ultrason. Ferroelectr. Freq. Control}, vol.~57, no.~11, Nov. 2010.

\bibitem{PCF}
J.~Camacho, M.~Parrilla, and C.~Fritsch, ``Phase coherence imaging,'' \emph{IEEE Trans. Ultrason. Ferroelectr. Freq. Control}, vol.~56, no.~5, 2009.

\bibitem{Chernyakova-Eldar_2018}
T.~Chernyakova \emph{et~al.}, ``Fourier-domain beamforming and structure-based reconstruction for plane-wave imaging,'' \emph{IEEE Trans. Ultrason. Ferroelectr. Freq. Control}, vol.~65, no.~10, 2018.

\bibitem{REFOCUS}
R.~Ali, C.~D. Herickhoff, D.~Hyun, J.~J. Dahl, and N.~Bottenus, ``Extending retrospective encoding for robust recovery of the multistatic data set,'' \emph{IEEE Trans. Ultrason. Ferroelectr. Freq. Control}, vol.~67, no.~5, pp. 943--956, 2020.

\bibitem{khan_real-time_2021}
C.~Khan, K.~Dei, S.~Schlunk, K.~Ozgun, and B.~Byram, ``A real-time, {GPU}-based implementation of aperture domain model image reconstruction,'' \emph{IEEE Trans. Ultrason. Ferroelectr. Freq. Control}, vol.~68, no.~6, pp. 2101--2116, Jun. 2021.

\bibitem{laroche_fast_2021}
N.~Laroche, S.~Bourguignon, J.~Idier, E.~Carcreff, and A.~Duclos, ``Fast non-stationary deconvolution of ultrasonic beamformed images for nondestructive testing,'' \emph{IEEE Trans. Comput. Imaging}, vol.~7, 2021.

\bibitem{IPB_Ozkan}
E.~Ozkan, V.~Vishnevsky, and O.~Goksel, ``Inverse problem of ultrasound beamforming with sparsity constraints and regularization,'' \emph{IEEE Trans. Ultrason. Ferroelectr. Freq. Control}, vol.~65, no.~3, pp. 356--365, 2018.

\bibitem{RED_USIPB}
S.~Goudarzi, A.~Basarab, and H.~Rivaz, ``Inverse problem of ultrasound beamforming with denoising-based regularized solutions,'' \emph{IEEE Trans. Ultrason. Ferroelectr. Freq. Control}, vol.~69, no.~10, 2022.

\bibitem{MNV2}
S.~Goudarzi, A.~Asif, and H.~Rivaz, ``Ultrasound beamforming using {MobileNetV2},'' in \emph{IEEE IUS}, 2020, pp. 1--4.

\bibitem{luijten_adaptive_2020}
B.~Luijten \emph{et~al.}, ``\BIBforeignlanguage{en}{Adaptive ultrasound beamforming using deep learning},'' \emph{\BIBforeignlanguage{en}{IEEE Trans. Med. Imaging}}, vol.~39, no.~12, Dec. 2020.

\bibitem{Hyun19}
D.~Hyun, L.~L. Brickson, K.~T. Looby, and J.~J. Dahl, ``Beamforming and speckle reduction using neural networks,'' \emph{IEEE Trans. Ultrason. Ferroelectr. Freq. Control}, vol.~66, no.~5, pp. 898--910, 2019.

\bibitem{perdios_cnn-based_2022}
D.~Perdios, M.~Vonlanthen, F.~Martinez, M.~Arditi, and J.-P. Thiran, ``{NN}-based image reconstruction method for ultrafast ultrasound imaging,'' \emph{IEEE Trans. Ultrason. Ferroelectr. Freq. Control}, vol.~69, no.~4, pp. 1154--1168, Apr. 2022.

\bibitem{Chennakeshava:ius2020}
N.~Chennakeshava, B.~Luijten, O.~Drori, M.~Mischi, Y.~C. Eldar, and R.~J.~G. van Sloun, ``High resolution plane wave compounding through deep proximal learning,'' in \emph{IEEE IUS}, 2020, pp. 1--4.

\bibitem{ZHANG2021}
J.~Zhang, Q.~He, Y.~Xiao, H.~Zheng, C.~Wang, and J.~Luo, ``Ultrasound image reconstruction from plane wave radio-frequency data by self-supervised deep neural network,'' \emph{Med. Image Anal.}, vol.~70, 2021.

\bibitem{van_sloun_deep_2020}
R.~J.~G. van Sloun, R.~Cohen, and Y.~C. Eldar, ``Deep learning in ultrasound imaging,'' \emph{Proc. IEEE}, vol. 108, no.~1, pp. 11--29, Jan. 2020.

\bibitem{ho_denoising_2020}
J.~Ho, A.~Jain, and P.~Abbeel, ``Denoising diffusion probabilistic models,'' \emph{NeurIPS}, vol.~33, pp. 6840--6851, 2020.

\bibitem{nichol_improved_2021}
A.~Q. Nichol and P.~Dhariwal, ``Improved denoising diffusion probabilistic models,'' in \emph{ICML}.\hskip 1em plus 0.5em minus 0.4em\relax PMLR, 2021, pp. 8162--8171.

\bibitem{dhariwal_diffusion_2021}
P.~Dhariwal and A.~Nichol, ``Diffusion models beat {GANs} on image synthesis,'' \emph{NeurIPS}, vol.~34, pp. 8780--8794, 2021.

\bibitem{kawar_denoising_2022}
B.~Kawar, M.~Elad, S.~Ermon, and J.~Song, ``Denoising diffusion restoration models,'' \emph{NeurIPS}, vol.~35, pp. 23\,593--23\,606, 2022.

\bibitem{song_solving_2022}
Y.~Song, L.~Shen, L.~Xing, and S.~Ermon, ``Solving inverse problems in medical imaging with score-based generative models,'' in \emph{ICLR}, 2022.

\bibitem{chung2022scoreMRI}
H.~Chung and J.~C. Ye, ``Score-based diffusion models for accelerated {MRI},'' \emph{Med. Image Anal.}, vol.~80, p. 102479, 2022.

\bibitem{DenoDDPM}
H.~Asgariandehkordi, S.~Goudarzi, A.~Basarab, and H.~Rivaz, ``Deep ultrasound denoising using diffusion probabilistic models,'' \emph{arXiv:2306.07440}, Jun. 2023.

\bibitem{dehaze}
T.~S. Stevens, F.~C. Meral, J.~Yu, I.~Z. Apostolakis, J.-L. Robert, and R.~J. van Sloun, ``Dehazing ultrasound using diffusion models,'' \emph{arXiv:2307.11204}, Jul. 2023.

\bibitem{horwitz2022conffusion}
E.~Horwitz and Y.~Hoshen, ``Conffusion: Confidence intervals for diffusion models,'' \emph{arXiv:2211.09795}, Nov. 2022.

\bibitem{zhang2023}
Y.~Zhang, C.~Huneau, J.~Idier, and D.~Mateus, ``Ultrasound image reconstruction with denoising diffusion restoration models,'' \emph{arXiv:2307.15990}, Jul. 2023.

\bibitem{iMAP}
T.~Chernyakova, D.~Cohen, M.~Shoham, and Y.~C. Eldar, ``{iMAP} beamforming for high-quality high frame rate imaging,'' \emph{IEEE Trans. Ultrason. Ferroelectr. Freq. Control}, vol.~66, no.~12, 2019.

\bibitem{quaegebeur_correlation-based_2012}
N.~Quaegebeur and P.~Masson, ``Correlation-based imaging technique using ultrasonic transmit–receive array for {Non}-{Destructive} {Evaluation},'' \emph{Ultrasonics}, vol.~52, no.~8, pp. 1056--1064, Dec. 2012.

\bibitem{DDIM}
J.~Song, C.~Meng, and S.~Ermon, ``Denoising diffusion implicit models,'' in \emph{ICLR}, 2021.

\bibitem{ILSVRC15}
O.~Russakovsky \emph{et~al.}, ``Imagenet large scale visual recognition challenge,'' \emph{Int. J. Comput. Vis.}, vol. 115, no.~3, pp. 211--252, 2015.

\bibitem{PICMUS}
H.~Liebgott, A.~Rodriguez-Molares, F.~Cervenansky, J.~Jensen, and O.~Bernard, ``Plane-wave imaging challenge in medical ultrasound,'' in \emph{IEEE IUS}, 2016.

\bibitem{FieII_1}
J.~Jensen and N.~Svendsen, ``Calculation of pressure fields from arbitrarily shaped, apodized, and excited ultrasound transducers,'' \emph{IEEE Trans. Ultrason. Ferroelectr. Freq. Control}, vol.~39, no.~2, pp. 262--267, 1992.

\bibitem{FieII_2}
J.~Jensen, ``Field: A program for simulating ultrasound systems,'' \emph{Med. Biol. Eng. Comput.}, vol.~34, no. sup. 1, pp. 351--353, 1997.

\bibitem{gCNR}
A.~Rodriguez-Molares \emph{et~al.}, ``The generalized contrast-to-noise ratio: A formal definition for lesion detectability,'' \emph{IEEE Trans. Ultrason. Ferroelectr. Freq. Control}, vol.~67, no.~4, pp. 745--759, 2020.

\bibitem{CUBDL}
D.~Hyun \emph{et~al.}, ``Deep learning for ultrasound image formation: {CUBDL} evaluation framework and open datasets,'' \emph{IEEE Trans. Ultrason. Ferroelectr. Freq. Control}, vol.~68, no.~12, pp. 3466--3483, 2021.

\bibitem{usDespeckle2022Lee}
H.~Lee, M.~H. Lee, S.~Youn, K.~Lee, H.~M. Lew, and J.~Y. Hwang, ``Speckle reduction via deep content-aware image prior for precise breast tumor segmentation in an ultrasound image,'' \emph{IEEE Trans. Ultrason. Ferroelectr. Freq. Control}, vol.~69, no.~9, pp. 2638--2650, 2022.

\bibitem{speckleModel2007Ng}
J.~Ng, R.~Prager, N.~Kingsbury, G.~Treece, and A.~Gee, ``Wavelet restoration of medical pulse-echo ultrasound images in an {EM} framework,'' \emph{IEEE Trans. Ultrason. Ferroelectr. Freq. Control}, vol.~54, no.~3, 2007.

\end{thebibliography}


\end{document}